# Multilevel profiling of situation and dialogue-based deep networks for movie genre classification using movie trailers

Dinesh Kumar Vishwakarma, Mayank Jindal, Ayush Mittal, Aditya Sharma
Biometric Research Laboratory, Department of Information Technology, Delhi Technological University, Delhi, India

**Abstract**

Automated movie genre classification has emerged as an active and essential area of research and exploration. Short duration movie trailers provide useful insights about the movie as video content consists of the cognitive and the affective level features. Previous approaches were focused upon either cognitive or affective content analysis. In this paper, we propose a novel multi-modality: situation, dialogue, and metadata-based movie genre classification framework that takes both cognition and affect-based features into consideration. A pre-features fusion-based framework that takes into account: situation-based features from a regular snapshot of a trailer that includes nouns and verbs providing the useful affect-based mapping with the corresponding genres, dialogue (speech) based feature from audio, metadata which together provides the relevant information for cognitive and affect based video analysis. We also develop the English movie trailer dataset (EMTD), which contains 2000 Hollywood movie trailers belonging to five popular genres: Action, Romance, Comedy, Horror, and Science Fiction, and perform cross-validation on the standard LMTD-9 dataset for validating the proposed framework. The results demonstrate that the proposed methodology for movie genre classification has performed excellently as depicted by the F1 scores, precision, recall, and area under the precision-recall curves.

1. **Introduction**

Movies are a great source of amusement for the audience, impacting society in numerous ways. Identifying the genre of a movie manually may vary due to an individual's taste. Hence, automated movie genre prediction is an active area of research and exploration. Movie trailers are becoming a useful source for predicting the genres of the movie. They provide useful insights into the movie in a very short duration of time. Movie trailers consist of two types of content: cognitive content and affective content.

Cognitive content describes the composition of the events, objects, and persons in a particular video frame of the movie trailer, while Affective content describes the types of psychological features such as feelings or emotions in a movie trailer [1]. Examples of cognitive content comprise a playground, a building, a man, a dog, etc. Examples of affective content are feelings/emotions such as happiness, sadness, anger, etc. Both the cognitive and affect-based content provide prominent features for predicting the genres of the movie.

In this paper, we propose a novel multi-modality situation, dialogue, and metadata-based movie genre classification framework, which aims to predict movie genres using video, audio and metadata (plot/description) content of movie trailers. Our novel framework focuses on extracting both the cognitive and affective features from the movie trailer. For achieving this, a sentence (generated from situations) composed of relevant nouns and verbs is extracted from the video frame. Nouns give the relevant information about the cognitive content of the trailers, and verbs provide useful affect-based mapping with the corresponding genres. For example, the verbs such as laughing, giggling, tickling, etc. provide an affect-based mapping with the 'comedy' genre. The verbs such as attacking, beating, hitting, etc. provide an affect-based mapping with the 'action' genre. Along with situations, dialogue and metadata-based features additionally contribute to cognitive and affective content as they include event descriptions (cognitive content) and psychological features (affective content).

Just like the standard machine learning process, the work is carried out in multiple phases. The 1st phase is the dataset generation phase, where we generate the EMTD, which contains 2000 Hollywood movie trailers belonging to 5 popular genres: Action, Romance, Comedy, Horror, and Science Fiction. The 2nd phase involves pre-processing of video trailers where all repeated frames are removed and resized. The sentences containing important nouns and verbs are extracted from the useful frames. We also prepare the audio transcripts of movie trailers to get dialogues from trailers. In the 3rd phase, we design and train the proposed architecture, which extracts and learn the important features from the trailers. Finally, in the 4th phase, the performance of our proposed architecture is evaluated using the Area under the Precision-Recall Curve (AU (PRC)) metric. The following are the significant contributions of our work:

- We propose a novel EMTD (English Movie Trailer Dataset) containing Hollywood movie trailers of the English language belonging to five popular and distinct genres: Action, Romance, Comedy, Horror, and Science Fiction.

- This work proposes a novel approach to predict movie genres using cognitive and affect-based features. None of the previous literature has focused on a combination of dialogue, situation, and metadata-based features extracted from the movie trailers to the best of our knowledge. Hence, we perform: situation-based analysis using nouns and verbs, dialogue-based analysis using speech recognition, and metadata-based analysis with metadata available with trailers.
- The proposed architecture is also evaluated by performing cross-dataset testing on the standard LMTD-9 [2] dataset. The results show that the proposed architecture has performed excellently and demonstrates the superior performance of the framework.

The remaining portion of the paper is organized as: In Section 2, the past literature on movie genre classification is reviewed, and the motivation behind the proposed work is highlighted. In Section 3, we discuss the proposed EMTD. In Section 4, we provide a detailed description of the proposed architecture. In Section 5, we evaluate the performance of the proposed framework and validate it against two different datasets. The paper is concluded in Section 6.

**2. Background and Related Work**

This section discusses the past methodologies for movie genre classification and the motivations behind our study. Video content is majorly partitioned to (1) Video frames (Images) and (2) Audio (Speech {dialogues} + Non speech {vocals}). To analyze the video content, various studies are done in the past, focusing chiefly upon cognitive [3]–[7] or affective [8] levels individually. For a more effective study, both levels need to be taken into account to perform better in a genre classification task.

In the past studies, many cognition-based approaches have been proposed based upon low-level features, including visual disturbances, average shot length, gradual change in light intensity in video frames, and peaks in audio waveform [3], to capture scene components [4]. Other features used for cognitive classification include RGB colors in frames [6], film shots [7], shot length [9], type of background in scenes (dark/non-dark) [6], etc. Similarly, some approaches are proposed for only affective analysis [8].

A movie can have multiple genres depicting a lot of information to viewers thus also serve as a task to recommend a movie to a viewer. Jain et al. [5] used 4 video features (shot length, motion, color dominance, lighting key) and 5 audio features to classify movie clips using only 200 training samples. They used complete movie clips to predict genres. However, the study uses only 200 training samples for training their model. Accordingly, the accuracy reported by them might be due to over-fitting. Also, the study focused only on single-label classification. Huang et al. [4] proposed the Self Adaptive Harmony Search algorithm with 7 stacked SVM's that used both audio and visual features (about 277 features in total) on a 223 sized dataset. Ertugrul et al. [10] used low-level features, including the movies' plot, by breaking the plot into sentences and classifying sentences into genres and taking the final genre to be one with maximum occurrence. Pais et al. [11] proposed to fuse image-text features by relying on some important words from the overall synopsis and performed movie genre classification based on those features. The model was tested on a set of 107 movie trailers. Shahin et al. [12] used movie plots and quotes and proposed Hierarchical attention networks to classify genres. Similarly, Kumar et al. [13] proposed to use movie plots to classify genre using hash vectorization by focusing on reducing overall time complexity. The above-mentioned studies rely on low-level features and do not capture any high-level features from movie trailers, thus cannot be relied upon for a good level recognition system.

From more recent studies, many researchers used deep networks for movie genre classification tasks. Shambharkar et al. [14] proposed a single label 3D CNN-based architecture to seize the spatial and temporal features. Though spatial and temporal features are captured in this, the model is not robust due to single-label classification. Some researchers have worked on movie posters to classify movie genres. Chu et al. [15] formulated a deep neural network to facilitate object detection and visual appearances. Though work captured a lot of information from posters, the poster itself is not enough to completely describe a movie. Simoes et al. [16] proposed a CNN-Motion that included scene histograms provided by the unsupervised clustering algorithm, weighted genre predictions for each trailer, along with some low-level video features. This provided a major group of features from a video but lacked some affective and cognitive-based features to classify genre.

Thus, from the past literature, it is evident that major information should be extracted from the video trailers for cognitive as well as affective study. So, our motivation behind the work is to device an approach relying on both levels of video content analysis as in [1]. We believe that the proposed architecture and the model are novel and robust and can be used in the future for various research perspectives.

## 3. EMTD Dataset

The datasets in previous literature lack the uniform composition of movie genres. Hence, we propose an EMTD (English Movie Trailer Dataset) consisting of around 2000 unique Hollywood movie trailers downloaded from IMDB[1]. EMTD contains 2000 unique trailers of 5 genres namely: action, comedy, horror, romance, science fiction. The dataset is extracted from IMDB by web scrapping procedure as follows: (1) fetch the list of movie titles available on IMDB (with at least 1 genre common to one mentioned above), (2) scrape metadata corresponding to each movie title including trailer link to download, and (3) download the trailers (.mp4) corresponding to the link into a folder, and list down all the information/meta-data about the movie including trailer name, descriptions, plot, keywords, and genres in the form of a CSV file. In this work, the dataset is partitioned into train set (1700 trailers), validation set (300 trailers) as shown in Table 1.

The study is conducted with the above genres only because mostly these genres are observed in the movies. We also want to explore the performance of our architecture first on a small set of genres, so we go for choosing only 5 genres instead of going towards a broad set of genres.

**Table 1: Dataset Composition**

| Genres / No. of trailers | Action | Comedy | Horror | Romance | Science Fiction |
|---|---|---|---|---|---|
| **Train set (1700)** | 671 | 637 | 422 | 388 | 555 |
| **Validation set (300)** | 126 | 135 | 49 | 67 | 99 |

## 4. Proposed Methodology

We propose a Multi-modality-based $M_{SD}$ framework (Pre-fused feature model) to perform the multi-label movie genre classification task. The Proposed model framework uses textual features i.e., situation-based features from video frames, dialogues extracted from audio, and meta-data from the dataset, as shown in Fig. 1. Majorly our framework is focused upon three features from movie trailers as also depicted in Fig. 1 (1) Descriptions, (2) Dialogues and (3) Situations (from frames). In the subsequent sections, the framework is discussed in detail.

### 4.1 Descriptions

Movie plot/descriptions are an important feature to describe a movie. In most cases, the plot mentioned for a movie being released is either too short or not mentioned in some cases.

---
[1] www.imdb.com

Considering this, we choose to use the descriptions concatenated with the dialogues extracted from movie trailers to finally predict the movie genre, as discussed in Section 4.2 in detail. Descriptions are fetched from the IMDB website as metadata as already mentioned in Section 3.

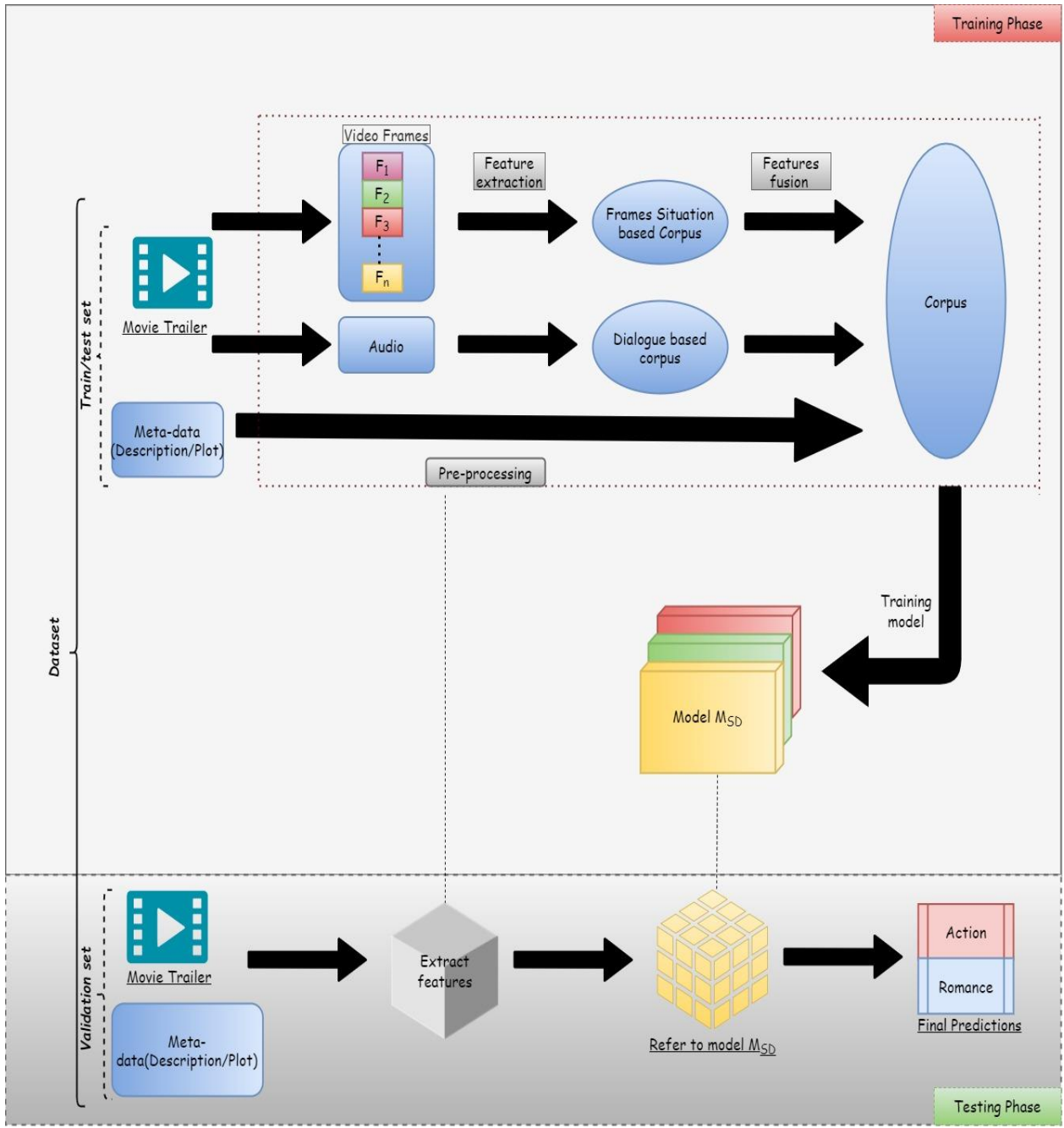

Fig. 1: Pipeline of the framework

### 4.2 Dialogue

In this section, we propose an architecture to process a list of dialogues from the trailer's audio (concatenated description/plot to dialogues) to predict movie genres. Significant steps for this stream include: (1) Extract speech (dialogue) from movie trailer and (2) Design a model to predict genres on the basis of speech and metadata.

**4.2.1 Data pre-processing**

The audio files in (.wav) format are extracted from the (.mp4) video trailers. Next, the audio file is split into small audio clips and converted to dialogues as proposed in [17]. All the text is collected to form an input corpus. Description/plot (if available in metadata) is also merged to this corpus. Our study is targeted for the English language trailers only. Just like movie plots, the speech extracted from the trailers can work as a supplement to our text corpus, which can help in better understanding of the relation between the text context and the genre of the movie. After generating the corpus comprised of a single record for each trailer in our training/testing phase the following pre-processing steps were conducted: converting all the text to lowercase, eliminating digits, punctuations, stop-words, and web-links. Text obtained above is used to feed as an input to the model/pre-trained model for training/testing.

**Table 2: Abbreviations with their meanings**

| Abbreviations | Meanings |
|---|---|
| EMTD | English Movie Trailer Dataset |
| LMTD-9 | Labeled Movie Trailer Dataset – 9 |
| $M_S$ | Situation based Model |
| $M_D$ | Dialogue based Model |
| $M_{SD}$ | Situation and Dialogue based Model |
| CNN | Convolution Neural Network |
| ANN | Artificial Neural Network |
| AU (PRC) | Area under the Precision–Recall curve |
| ADAM | Adaptive Moment Estimation |
| Bi-LSTM | Bidirectional Long Short-Term Memory |
| TF-IDF | Term Frequency - Inverse Document Frequency |
| E-Bi LSTM | Embedding-Bidirectional LSTM |
| ECnet | Embedding – Convolution network |
| TFAnet | Term Frequency Artificial Neural Network |

**4.2.2 Feature Extraction (Dialogue)**

The large audio file is split into smaller chunks based on silence. These chunks are then processed, and the final text is obtained by concatenating all processed chunks. An audio file containing a sequence of 'n' chunks can be represented as $A = \{c_1, c_2, c_3, \ldots c_n\}$, where $c_i$ ($1 \leq i \leq n$) is the i<sup>th</sup> chunk in audio – A. Let $f(A)$ be the function that takes audio as an input and generates dialogues from it defined as Eq. (1) :

$$f(A) = f(c_1, c_2, c_3, \ldots c_n) = concat(\widehat{W}_{c_i}), 1 \leq i \leq n \tag{1}$$

$\widehat{W}_{c_i}$ is the maximum a posteriori probability $P(W_{c_i}|X_{c_i})$ of predicting the optimal word sequence $W_{c_i}$, given the speech signal $X_{c_i}$ for each audio chunk $c_i$.

$$\widehat{W}_{c_i} = \arg\max_w P(W_{c_i}|X_{c_i}) \qquad (2)$$

Where $P(W_{c_i}|X_{c_i})$ is defined using Bayes' Rule as Eq. (3).

$$P(W_{c_i}|X_{c_i}) = \frac{p(X_{c_i}|W_{c_i})P(W_{c_i})}{p(X_{c_i})} \qquad (3)$$

In case of conditional independence, Eq. (3) can be written as Eq. (4).

$$P(W_{c_i}|X_{c_i}) \propto p(X_{c_i}|W_{c_i})P(W_{c_i}) \qquad (4)$$

Substituting the value of conditional independence in Eq. (2), it can be written as Eq. (5).

$$\widehat{W}_{c_i} = \arg\max_w p(X_{c_i}|W_{c_i})P(W_{c_i}) \qquad (5)$$

$p(X_{c_i}|W_{c_i})$ is the acoustic model likelihood and $P(W_{c_i})$ is the language model probability as mentioned in [17]. For each audio chunk $c_i$, when the time sequence is expanded and the observations $x_{c_{i_t}}$ are assumed to be generated by hidden Markov models (HMMs) with hidden states $\alpha_{c_{i_t}}$, and can be written as Eq. (6).

$$\widehat{W}_{c_i} = \arg\max_w P(W_{c_i}) \sum_\alpha \prod_{t=1}^T p\left(x_{c_{i_t}}|\alpha_{c_{i_t}}\right) P\left(\alpha_{c_{i_t}}|\alpha_{c_{i_{t-1}}}\right) \qquad (6)$$

where $\alpha$ belongs to the set of all possible state sequences for the transcription $W_{c_i}$. Substituting $\widehat{W}_{c_i}$ from Eq. (6) in Eq. (1), we get Eq. (7).

$$f(A) = concat\left(\arg\max_w P(W_{c_i}) \sum_\alpha \prod_{t=1}^T p\left(x_{c_{i_t}}|\alpha_{c_{i_t}}\right) P\left(\alpha_{c_{i_t}}|\alpha_{c_{i_{t-1}}}\right)\right), 1 \leq i \leq n \qquad (7)$$

Eq. (7) represents the final dialogues generated from audio.

### 4.2.3 ECnet (Embedding – Convolution network)

To build cognitive-based genre detection architecture, the crucial features of the trailer in the form of a text corpus needs to be learned by a model. This can be achieved by using a combination of Embedding and CNN (Convolution Neural Network) layers. The layers of the multi-label classification network is depicted in Table 3. Embedding is one of the popular techniques used in NLP problems for converting words into mathematical representation in the form of numerical vectors.

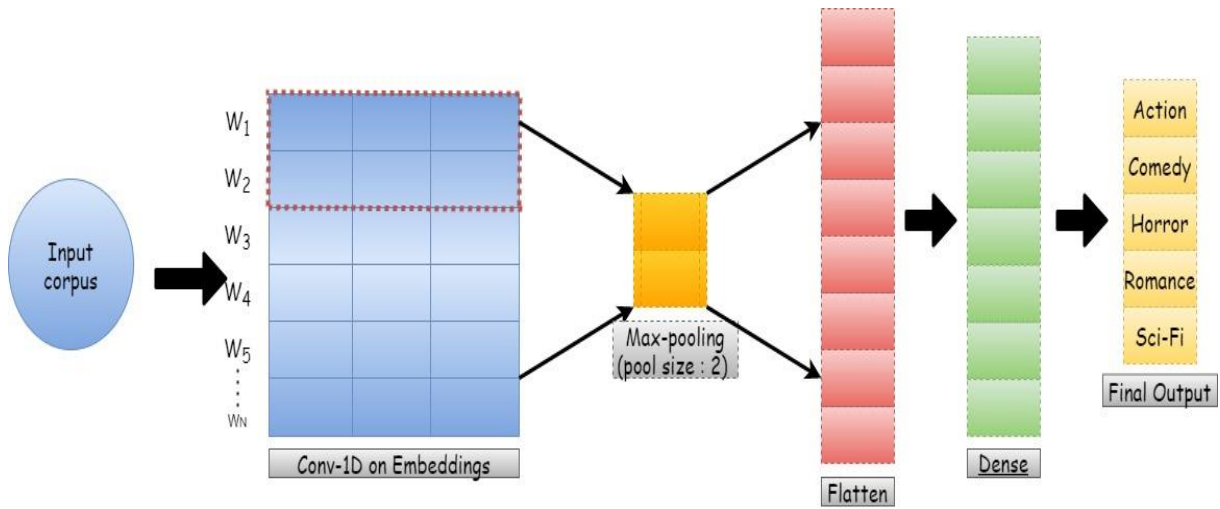

Fig. 2: ECnet architecture

Before actually sending input to the architecture, the vocabulary needs to be designed and the size of a corpus for each data point needs to be fixed. A vocabulary of size 10,395 words is designed and the maximum length of the number of words in each corpus is set to be the length of the longest sentence in our training corpus, which is 330 in our case. If the number of words in a corpus is less than the max length, the corpus is padded with 0's. For a 2-3-minute movie trailer, 330 words are found to be sufficient as in some parts of the trailer there may be no speech (only vocals might be present).

Now for each corpus in the input data, we are having an input of shape (330,) (330 is the number of words in each data point), which is fed to the first layer of our architecture as in Fig. 2, i.e., embedding layer. The embedding layer gives an output of dimension (330, 64,) as the length of embedding for each word is taken to be 64 in our proposed architecture.

**Table 3: Parameters of ECnet architecture**

| Layer Name: Type | Description | Input Shape | Output Shape | Parameters # |
|---|---|---|---|---|
| **embedding: Embedding** | Embedding Layer with input length = 330 | (330,) | (330, 64,) | 665280 |
| **conv1d: Conv1D** | Conv1D Layer with 64 filters | (330, 64,) | (330, 64,) | 12352 |
| **max_pooling1d: MaxPooling1D** | MaxPooling1D Layer with pool size = 2 | (330, 64,) | (165, 64,) | 0 |
| **flatten: Flatten** | Flatten Layer | (165, 64,) | (10560,) | 0 |
| **dense_1: Dense** | Dense Layer with 32 neurons | (10560,) | (32,) | 337952 |
| **dense_2: Dense** | Dense Layer with 5 neurons | (32,) | (5,) | 165 |

After the embedding layer, a 1-D convolution layer is fed with the output of the embedding layer. Again, the convolution layer gives an output shape of (330, 64,). To get the same output, we apply the padding uniformly to the input of the convolution layer. Next, a max-pooling layer is used to reduce the dimension of data from (330, 64,) to (165, 64,). The architecture is followed by a flatten layer to transform the two-dimensional data into one-dimensional data, to further send the output to a dense layer.

As depicted in Table 3, the flatten layer gives an output of shape (10560,) which is fed to a dense layer as input and giving an output shape of (32,). Finally, the final dense layer is applied to the architecture returning the output shape of (5,) denoting our five genres. In our architecture's final dense layer, we use "sigmoid" as an activation function best suited for our multi-label classification problem.

### 4.3 Situation

This section includes the work we proposed on visual features from movie trailers. Primary steps for this stream include: (1) fetch video frames from the trailer, (2) extract situations from the frames and (3) build architecture to finally classify the trailers into genres.

A novel situation-based video analysis model is proposed by extracting the situations and events based on each frame extracted from the video for visual features. Thus, a corpus is created to train/test the model by collecting them together.

To the best of our knowledge, we are proposing a novel framework by fusing the situation, event, and dialogue analysis for genre classification. More details about the framework are described in the below sections.

#### 4.3.1 Frame Extraction from Video

Consider a movie trailer video $V_t$, it can be expressed as a sequence of frames and represented as Eq. (8).

$$V^t = \{F_1^t, F_2^t, F_3^t, \ldots\ldots F_n^t\}, \tag{8}$$

where $n$ is the total number of frames in trailer $t$ and $F_i$ is the $i^{th}$ frame of trailer $t$.

After various experimentation using some subset of movie trailers, it is found that taking every $10^{th}$ the frame is beneficial to avoid redundancy in frames (consecutive frames from a video appear to be similar). Hence, after discarding the redundant frames, the final video frames considered can be expressed as Eq. (9):

$$(V^t)_f = \{F_1^t, F_{21}^t, F_{31}^t, \ldots\ldots F_m^t\} \tag{9}$$

In the subsequent sections, we consider these frames for every trailer.

### 4.3.2 Feature Extraction (Situation)

In this work, we aim to extract a mixture of cognitive and affective features from the video frames. There has been a lot of work on identifying the objects from the frames, as discussed in 2, which provides only cognitive features. For situation analysis, we extracted features from Grounded Situation Recognition [18] to identify both cognitive and affective features. Situation-based features describing the frame are extracted for each frame from the set $(V^t)_f$, which contains verbs depicting affect-based features, nouns (agent, item, part, place, stage, tool) depicting cognitive ones, and the verb definitions as depicted in Fig. 3.

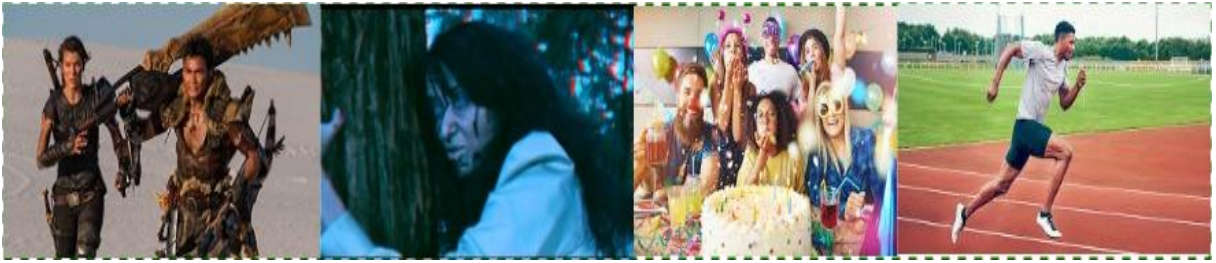

Fig. 3: Situations for above frames: (a) the soldiers marches in outdoor. (b) haunted lady with blood on face standing along tree (c) people celebrates birthday at a room. (d) a man sprints at a racetrack

To extract features from frames of trailer, the work proposed in [18] to produce situations from images is used. Every image can be expressed in a combination of associated verb and semantic roles present in an image. For every verb, a semantic role can be formed describing the situation like, for eating it can be $Role_{crying} = \{Agent\ (man), Place(road)\}$, where $Role_{crying}$ is semantic role for verb crying. Hence, a situation in image $I$ can be expressed as Eq. (10).

$$S = \{v,\ A_{(v,I)}\}, v\ is\ verb\ expressed\ in\ situation\ S \tag{10}$$

where, $A_{(v,I)} = \{(r_i, n_i), where\ r_i \in Role_v, n_i \in Nouns\ 1 \leq i \leq |Role_v|\ \}$

And the probability that situation S belongs to an image I can be denoted as in Eq. (11).

$$P(S/I, \alpha) = P(v, A_{(v,I)}, \alpha) \tag{11}$$

We can replace $A_{(v,I)}$ in Eq. (11) from Eq. (10). Hence, $P(S/I, \alpha)$ can further be expressed as in Eq. (12).

$$P(S/I,\alpha) = P\big(v,(r_1,n_1),(r_2,n_2),(r_3,n_3),\ldots\ldots,(r_{|Roles_v|},n_{|Role_v|})|I;\alpha\big) \quad (12)$$

$\alpha$ denotes the parameter for our neural; network. Now, we can define the semantic roles in an image in a particular order. Thus further, the Eq. (12) be reduced to Eq. (13).

$$P(S/I,\alpha) = P\big(v,n_1,n_2,n_3,\ldots\ldots,n_{|Role_v|}|I;\alpha\big) \quad (13)$$

Eq. (13) can be further simplified as Eq. (14).

$$P(S/I,\alpha) = P(v/I)\prod_{j=1}^{|Role_v|} P(n_j|v,n_1,n_2,n_3,\ldots\ldots,n_{j-1}|I;\alpha) \quad (14)$$

For a given particular image/frame, the situation having maximum value probability defined in Eq. (14) will be considered for that image.

The situation for every frame is extracted using the above method, $S^t = \{S_{F_1^t}, S_{F_{21}^t}, \ldots\ldots, S_{F_m^t}\}$, $F_i^t$ is defined in Section 4.3.1 where $S^t$ is the set of the situation extracted for every frame. Finally, the corpus situation text $C^t$ is formed by the concatenation of individual situations from frames as in Eq. (15).

$$C^t = concat(S_i), \text{ where } i \in \big[(V^t)_f\big] \quad (15)$$

Now the task is converted to a text classification task for which we propose the model architecture as discussed in upcoming sections. Before proceeding to the next step, text pre-processing is conducted: converting all the text to lowercase, eliminating digits, punctuations, and stop-words, as mentioned in Section 4.2.1. These same steps are performed in the testing procedure to predict the movie trailer genre.

**4.3.3 TFAnet (Term Frequency Artificial Neural Network)**

After extracting visual features, a robust architecture is required to classify the final genres for the trailers. This model is different from the model we proposed in the dialogue stream. Here, TFAnet (Term Frequency Artificial Neural Network) is proposed consisting of a deep network of dense and dropout layers as depicted in Fig. 4.

Before coming to the proposed architecture, we will discuss the text representation using TF-IDF in [19]. For this architecture, it is proposed to use in the word count in the corpus of each data point. Hence, we use the word count from the corpus as features for classifying the movie trailer genres. In order to get a large number of words included as features in our vocabulary set, trailers from a large range of released dates are used in our EMTD to get a huge corpus available with us while training the model. A combination of unigrams, bigrams and trigrams is used from our corpus as the features and TF-IDF (term frequency-inverse document frequency) algorithm represents our text in a numerical form. The total n-grams features taken are around 34,684. Now our text-based features are transformed into mathematical form, so next (artificial neural network) is trained to classify the genres of the trailer.

**Table 4: Parameters of TFAnet**

| Layer Name: Type | Description | Input Shape | Output Shape | Parameters # |
|---|---|---|---|---|
| **dense_1: Dense** | Dense Layer with 64 neurons | (34684,) | (64,) | 2219840 |
| **dropout_1: Dropout** | Dropout Layer with rate = 0.4 | (64,) | (64,) | 0 |
| **dense_2: Dense** | Dense Layer with 32 neurons | (64,) | (32,) | 2080 |
| **dropout_2: Dropout** | Dropout Layer with rate = 0.2 | (32,) | (32,) | 0 |
| **dense_3: Dense** | Dense Layer with 5 neurons | (32,) | (5,) | 165 |

The architecture of TFAnet (Term Frequency Artificial Neural Network) is depicted in Table 4. The input shape, as discussed above, is (34684,). This input is given to a dense layer, which gives an output of shape (64,). Then a dropout layer is applied to reduce over-fitting with a rate of 0.4. Again, a dense layer is applied, and we obtain an output of shape (32,), followed by a dropout layer with a rate of 0.2. Finally, a dense layer is applied, which gives an output of shape (5,) to finally predict five genres, with sigmoid as an activation function.

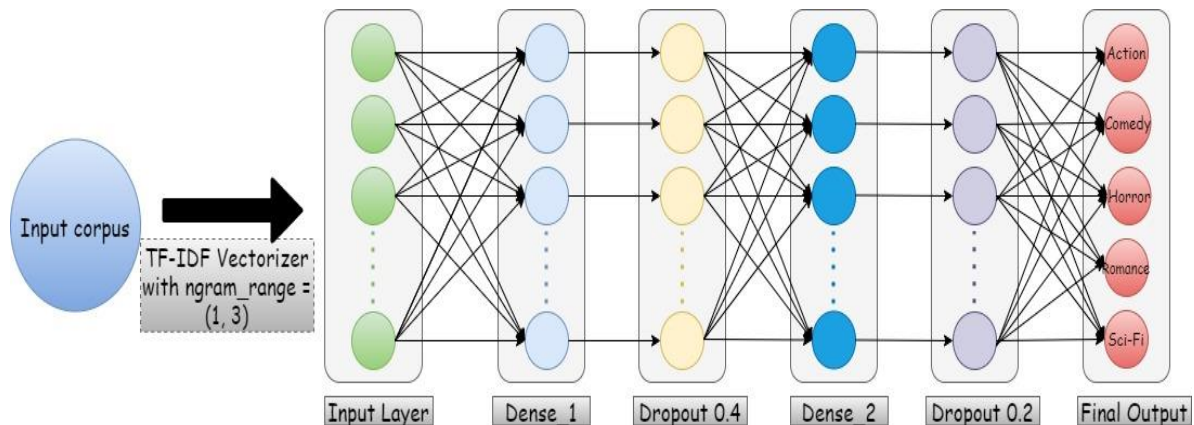
Fig. 4: TFAnet Architecture

The algorithm of the training phase of $M_{SD}$ model is written as Algorithm 1.

**Algorithm 1: Training phase of $M_{SD}$ model**

**INPUT:** Set of movie trailers along with meta-data for the training of $M_{SD}$
**OUTPUT:** Trained $M_{SD}$ with learned weights ready to predict multi-label genres.

    **START**
1    $C_F = \Phi$

```
2    for every trailer in training dataset:
3        Situation extraction
4        C_S = Φ
5        C_M = description of trailer from metadata
6            for every 10th frame in trailer:
7                resize frame to 299*299*3
8                extract situations from frame
9                append situation to corpus C_S
10       Dialogues extraction
11       A_d = fetch audio from trailer
12       C_D = get speech from A_d
13       Corpus fusion
14       C_SD = C_D + C_S + C_M
15       Append C_SD to C_F
16   design ECnet architecture
17   train M_SD using C_F
     END
```

$C_F$: list of input corpora
$C_S$: situation corpus for a trailer
$C_D$: dialogue for a trailer
$C_M$: description from metadata
$C_{SD}$: a concatenated corpus for a trailer
$A_d$: audio from a trailer

The process of the testing phase can be understood with Algorithm 2.

**Algorithm 2: Testing phase of $M_{SD}$ model**

**INPUT:** Set of movie trailers along with meta-data for the test set.
**OUTPUT:** Genres predicted to test set movie trailers.

```
    START
1   for every trailer in the test set
2       C_SD = extract features of the trailer using step 3 to 14 of algorithm 1
3       P_t = get a prediction from pre-trained model M_SD
4       output predicted genres
    END
```

$C_{SD}$: a concatenated corpus for a trailer
$P_t$: predictions corresponding to trailer t

**4.4 $M_{SD}$ Framework**

$M_{SD}$ framework is based on the pre-feature fusion of all 3 modalities as described in the previous sections. It uses a similar architecture as the one proposed in Section 4.2.3. Only a certain number of parameters are changed, like the size of vocabulary, which in this case is 14172, and the maximum length of the corpus for each data point is 1661. Algorithm1 and Algorithm2 depict the framework $M_{SD}$. The final feature fusion is expressed as Eq. (13).

$$F_{SD} = F_S \parallel F_D \parallel F_M \tag{13}$$

where $F_S$, $F_D$ and $F_M$ represent the situation, dialogue, and metadata features, respectively. $F_{SD}$ is obtained by concatenating $F_S$, $F_D$ and $F_M$ for each trailer.

## 5. Experiments

In this part, we will examine various model architectures on different modalities and pre-feature fused models. Later, we verify our work by validating it on the standard LMTD-9 dataset as well as on our proposed dataset. Finally, a comparative study is discussed to explore our model robustness. All the experiments are performed on GPU workstations with 128 GB DDR4 RAM and Nvidia Titan RTX (24 GB) GPU configuration.

### 5.1 Datasets

To verify our framework, we utilize our proposed dataset and standard LMTD-9 [2] dataset. Comprehensive details are mentioned as follows:

#### 5.1.1 English movie trailer dataset (EMTD)

EMTD: Our proposed dataset contains a separate training set of 1700 unique trailers and a validation set of 300 unique trailers, all taken from IMDB, as mentioned in Section 3.

#### 5.1.2 Labeled movie trailer dataset (LMTD-9)

LMTD [16], [20] is a multi-label large-scale movie trailer dataset including trailer link, trailer metadata, plot/summary, unique trailer id consisting of around 9k movie trailers belonging to 22 distinct labels/genres. For verification purposes, a validation set (subpart) of LMTD-9 [2] is used that only includes the Hollywood trailers released after 1980 and trailers specific to our genre list. The dataset contains varying length trailers with different video quality and aspect ratios.

### 5.2 Classification results on different models

In this section, we will discuss our experiments with different framework variations. We experimented with 3 different frameworks based on separate modalities and pre-fused features.

- **M$_S$ (Video frames analysis):** Model considering the only Situation based features from video frames.
- **M$_D$ (Dialogues-metadata analysis):** Model considering dialogues from audio and descriptions from metadata as features.
- **M$_{SD}$ (Multi-modality analysis):** Model considering situation-based features from video

frames, dialogues from audio and descriptions from metadata as features.

For $M_S$, we build an ECnet architecture using Keras framework in python as per the architecture defined in Section 4.2.3. The data is split into 85:15 ratios for training and testing. The model is trained on 50 epochs with a learning rate of 0.001 using Adam [21] optimizer to get a training accuracy of 83% and testing accuracy of 70% in 50 epochs.

For $M_D$, we build a TFAnet (Term Frequency Artificial Neural Network) architecture using Keras framework in python using the layers defined in Section 4.3.3. The data is split into 85:15 ratios for training and testing. The model is trained on 50 epochs with a learning rate of 0.001 using Adam [21] optimizer to get a training accuracy of 80% and testing accuracy of 60% in 50 epochs. After training, $M_S$ and $M_D$ are validated on both datasets.

Table 5: Classification results of $M_{SD}$ on EMTD and LMTD-9

| Model | | $M_{SD}$ | |
|---|---|---|---|
| Genres | Dataset | EMTD | LMTD-9 |
| Action | P | 0.88 | 0.89 |
|  | R | 0.89 | 0.89 |
|  | F1 | 0.88 | 0.89 |
| Romance | P | 0.79 | 0.71 |
|  | R | 0.76 | 0.77 |
|  | F1 | 0.77 | 0.73 |
| Horror | P | 0.88 | 0.89 |
|  | R | 0.81 | 0.87 |
|  | F1 | 0.84 | 0.88 |
| Sci-fi | P | 0.89 | 0.68 |
|  | R | 0.86 | 0.80 |
|  | F1 | 0.87 | 0.75 |
| Comedy | P | 0.86 | 0.85 |
|  | R | 0.85 | 0.85 |
|  | F1 | 0.85 | 0.84 |

The architecture proposed in Section 4.2.3 with pre-fused features is used for $M_{SD}$. However, the input corpus is slightly modified. The corpus defined in Section 4.4 is used for $M_{SD}$. Precision, Recall, and F1-score for $M_{SD}$ on LMTD-9 and EMTD is depicted in Table 5. However, the AU (PRC) comparison of $M_{SD}$ with $M_S$ and $M_D$ is discussed in the upcoming section.

Some variation can be seen among the performance of different genres. Most of the trailers belonging to major genres are being classified precisely (with an F1 score of 0.84 and above), which shows that the proposed model is performing well. The action genre was the best performing genre among five with an F1-score of 0.88 and 0.89 on EMTD and LMTD-9 respectively. The romance genre was seen to be the least performing genre among all genres in terms of the F1-score. It is observed that many romance genre trailers are being misclassified into comedy as both these genres are dominated by similar words like happy, smile, laugh, etc.

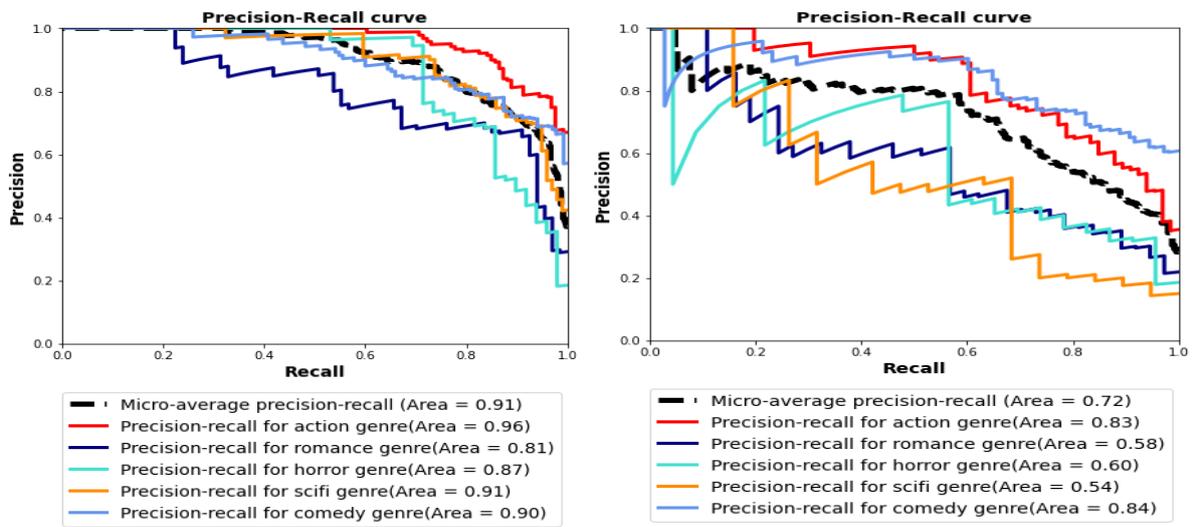

Fig. 5: Precision recall curve for $M_D$  A) EMTD    B) LMTD-9

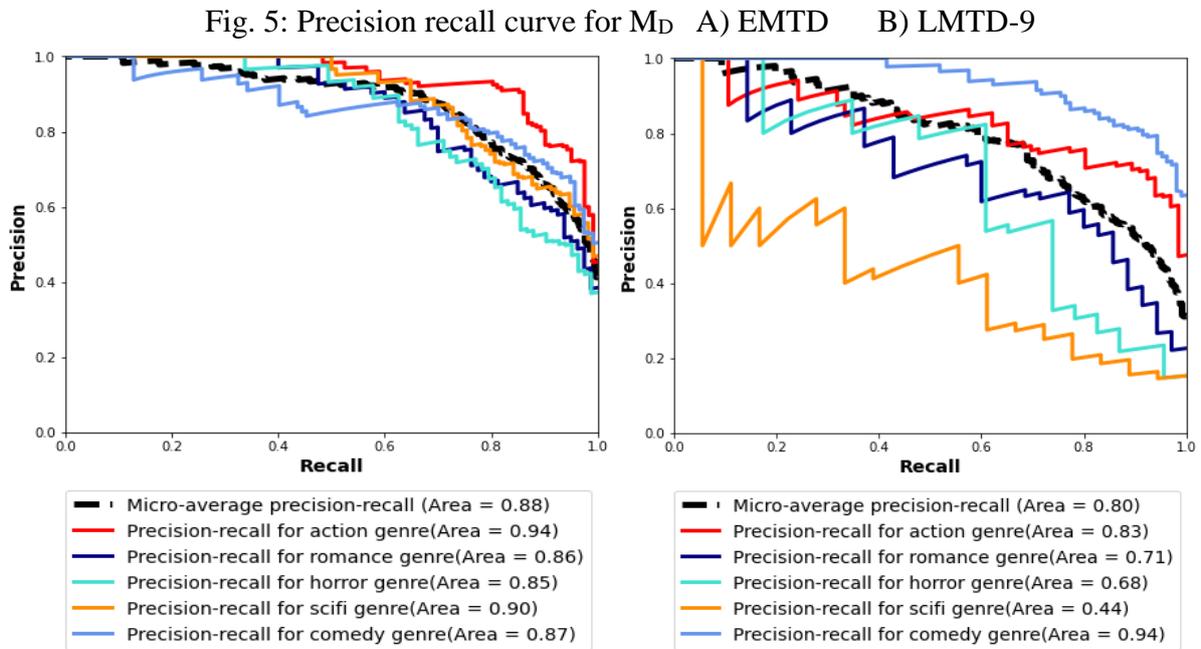

Fig. 6: Precision recall curve for $M_S$  A) EMTD    B) LMTD-9

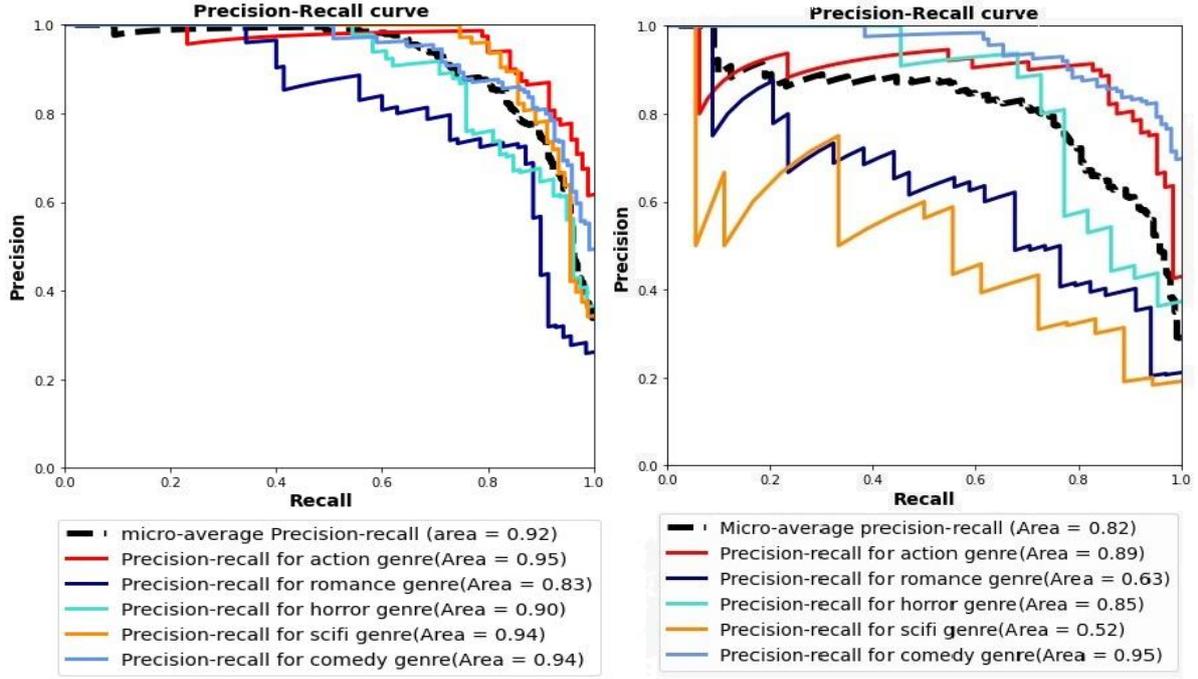

Fig. 7: Precision-Recall Curves $M_{SD}$  A) EMTD    B) LMTD-9

### 5.3 AU (PRC) Comparison

The AU (PRC) i.e., area under the precision-recall curve, is calculated to compare our classification results, as we are dealing with the multi-label classification problem. AU (PRC) measure helps to compare the actual performance of our model, compensating for the noise effect due to class imbalance in the multi-label dataset. The AU (PRC) curves are created for all 3 models on both datasets as depicted in Fig. 5, Fig. 6, and Fig. 7. On the validation set of EMTD, we found almost similar AU (PRC) values 92%, 91%, 88% on $M_{SD}$, $M_D$, and $M_S$, respectively. However, we found that our $M_{SD}$ gives the 82% AU (PRC) values on the LMTD-9 dataset, which is greater than the other two models i.e., 72% and 80% AU (PRC) of $M_D$ and $M_S$ respectively as in Table 6.

**Table 6: AU (PRC) on different models**

| Features | Model | Dataset | |
| --- | --- | --- | --- |
| | | EMTD | LMTD-9 |
| **Dialogue ($M_D$)** | E-Bi LSTM | 0.87 | 0.66 |
| | ECnet | 0.91 | 0.72 |
| **Situation ($M_S$)** | ECnet | 0.86 | 0.75 |
| | TFAnet | 0.88 | 0.80 |
| **Fused Features ($M_{SD}$)** | ECnet | **0.92** | **0.82** |

However, for overall comparison with some other models that we experimented with within our study, we mention their results in Table 6. For choosing the best architecture, the models are compared in terms of AU (PRC) on both of the validation datasets. The implementation of all the mentioned models helps us in deciding the best model for the fused features. Although $M_D$ has comparable AU (PRC) values with $M_{SD}$ on EMTD but on LMTD-9, $M_{SD}$ outperformed $M_D$. Similar is the case with $M_S$ on LMTD-9. While $M_{SD}$ performed simultaneously well on both datasets, which is not true in the case of $M_S$ and $M_D$ individually. So, by cross dataset validation $M_{SD}$ proves to be a more robust one. We conclude that the proposed $M_{SD}$ is the best performing model.

### 5.4 Baseline comparison

In this section, we validate the performance of our proposed model by performing the state of art comparison with the previous approaches for movie genre classification using the AU (PRC) metric for each genre separately as depicted in Table 7. All the results mentioned in Table 7 are shown up to two decimal places and are based on the standard LMTD-9 dataset except for Fish et. al. [22], whose results are based on MMX trailer-20 dataset. It does not consider the romance genre in its study. However, for the other genres, the difference in the AU (PRC) values of Fish et. al [22] and $M_{SD}$ is worth noting. $M_{SD}$ outperforms it by 20% on average. Low-level visual features based classification [23] is based on 24 low-level visual features, SAS-MC-v2 [24] uses only the synopsis for trailer classification, Fish et. al. [22] and CTT-MMC-TN [25] are based on high-level features. Comparing to low-level feature approaches [23], [24], $M_{SD}$ on an average outperforms by 10%, and by comparing from approaches using high-level features [22], [25], it outperforms by 8% on an average for each genre. It is also observed comedy genre performed well in most works as compared to the other four genres while science-fiction has relatively lower AU (PRC) values. This could be due to the unavailability of proper distinction in the science-fiction genre, as its features overlap with some other similar genres (like action).

**Table 7: Comparison of the proposed model with similar state-of-the-arts using AU (PRC)**

| S. No. | Methods | Action | Romance | Sci-fi | Horror | Comedy |
|---|---|---|---|---|---|---|
| 1 | Low-level visual features [23] | 0.85 | 0.46 | 0.19 | 0.42 | 0.87 |
| 2 | SAS-MC-v2 [24] | 0.68 | 0.48 | 0.52 | 0.58 | 0.74 |
| 3 | Fish et. al. [22] | 0.62 | ---- | 0.50 | 0.56 | 0.71 |
| 4 | CTT-MMC-TN [25] | 0.83 | 0.45 | 0.40 | 0.66 | 0.87 |
| 5 | $M_{SD}$ | **0.89** | **0.63** | **0.52** | **0.85** | **0.95** |

The comparative study demonstrates that the proposed model is robust as it outperforms existing approaches and gives excellent results. The better performance is due to the reason that the architecture proposed includes both cognitive and affective features, helping the model to learn substantial characteristics of each genre, hence predicting genres more precisely.

## 6 Conclusion

This work extends the idea of a novel holistic approach to the movie genre classification problem that includes affective and cognitive levels by considering multiple modalities, including situation from the frame, dialogues from speech, and meta-data (movie plot and description). We also built a Hollywood English movie trailers dataset EMTD that includes around 2000 trailers from 5 genres, namely action, comedy, horror, romance, science fiction, to pursue this study. We experimented with various model architectures as discussed in Section 5.2 and also validated our final framework on EMTD and on standard LMTD-9 [2] that achieves AU (PRC) values of 0.92 and 0.82 respectively. Our study's main aim is to build a robust framework to classify a movie genre from its short clip i.e., trailer. Although our study includes English speech as a feature, it can also be applied to some Non-English trailers. For Non-English ones, our model can incorporate the video features only, so on the basis of that, predictions can be made by our architecture.

For extension of our proposed model, background audio studies based on vocals can also be incorporated. Hence, in the future, we plan to build a framework considering background vocals in audio along with our current framework to better extract and use most features from movie trailers. We also plan to add some more genres to our study for multi-label classification.

## 7 References

[1] A. Hanjalic and L. Q. Xu, "Affective video content representation and modeling," *IEEE Trans. Multimed.*, vol. 7, no. 1, 2005.
[2] J. Wehrmann and R. C. Barros, "Convolutions through time for multi-label movie genre classification," in *Proceedings of the ACM Symposium on Applied Computing*, 2017, vol. Part F1280, pp. 114–119.
[3] Z. Rasheed, Y. Sheikh, and M. Shah, "On the use of computable features for film classification," *IEEE Trans. Circuits Syst. Video Technol.*, vol. 15, no. 1, pp. 52–64, Jan. 2005.
[4] L. H. Chen, Y. C. Lai, and H. Y. Mark Liao, "Movie scene segmentation using background information," *Pattern Recognit.*, vol. 41, no. 3, 2008.
[5] S. K. Jain and R. S. Jadon, "Movies genres classifier using neural network," 2009.
[6] L. Canini, S. Benini, and R. Leonardi, "Affective recommendation of movies based on selected connotative features," *IEEE Trans. Circuits Syst. Video Technol.*, vol. 23, no. 4, 2013.
[7] M. Xu, C. Xu, X. He, J. S. Jin, S. Luo, and Y. Rui, "Hierarchical affective content analysis in arousal and valence dimensions," *Signal Processing*, vol. 93, no. 8, 2013.
[8] A. Yadav and D. K. Vishwakarma, "A unified framework of deep networks for genre classification using movie trailer," *Appl. Soft Comput. J.*, vol. 96, 2020.
[9] K. Choroś, "Video genre classification based on length analysis of temporally aggregated video shots," in *Lecture Notes in Computer Science (including subseries Lecture Notes in Artificial Intelligence and Lecture Notes in Bioinformatics)*, 2018, vol. 11056 LNAI, pp. 509–518.

[10] A. M. Ertugrul and P. Karagoz, "Movie Genre Classification from Plot Summaries Using Bidirectional LSTM," in *Proceedings - 12th IEEE International Conference on Semantic Computing, ICSC 2018*, 2018, vol. 2018-January.
[11] G. Païs, P. Lambert, D. Beauchêne, F. Deloule, and B. Ionescu, "Animated movie genre detection using symbolic fusion of text and image descriptors," 2012.
[12] A. Shahin and A. Krzyżak, "Genre-ous: The Movie Genre Detector," in *Communications in Computer and Information Science*, 2020, vol. 1178 CCIS.
[13] N. Kumar, A. Harikrishnan, and R. Sridhar, "Hash Vectorizer Based Movie Genre Identification," in *Lecture Notes in Electrical Engineering*, 2020, vol. 605.
[14] P. G. Shambharkar, P. Thakur, S. Imadoddin, S. Chauhan, and M. N. Doja, "Genre Classification of Movie Trailers using 3D Convolutional Neural Networks," 2020.
[15] W. T. Chu and H. J. Guo, "Movie genre classification based on poster images with deep neural networks," 2017.
[16] G. S. Simões, J. Wehrmann, R. C. Barros, and D. D. Ruiz, "Movie genre classification with Convolutional Neural Networks," in *Proceedings of the International Joint Conference on Neural Networks*, 2016, vol. 2016-October.
[17] J. Li, L. Deng, R. Haeb-Umbach, and Y. Gong, "Chapter 2 - Fundamentals of speech recognition," in *Robust Automatic Speech Recognition*, J. Li, L. Deng, R. Haeb-Umbach, and Y. Gong, Eds. Oxford: Academic Press, 2016, pp. 9–40.
[18] S. Pratt, M. Yatskar, L. Weihs, A. Farhadi, and A. Kembhavi, "Grounded Situation Recognition," in *Computer Vision -- ECCV 2020*, 2020, pp. 314–332.
[19] B. Beel, Joeran and Langer, Stefan and Gipp, "TF-IDuF: A Novel Term-Weighting Sheme for User Modeling based on Users' Personal Document Collections," *Proc. iConference 2017*, 2017.
[20] J. Wehrmann, R. C. Barros, G. S. Simoes, T. S. Paula, and D. D. Ruiz, "(Deep) Learning from Frames," 2017.
[21] D. P. Kingma and J. L. Ba, "Adam: A method for stochastic optimization," 2015.
[22] E. Fish, A. Gilbert, and J. Weinbren, "Rethinking movie genre classification with fine-grained semantic clustering," *arXiv Prepr. arXiv2012.02639*, 2020.
[23] F. Álvarez, F. Sánchez, G. Hernández-Peñaloza, D. Jiménez, J. M. Menéndez, and G. Cisneros, "On the influence of low-level visual features in film classification," *PLoS One*, vol. 14, no. 2, 2019.
[24] J. Wehrmann, M. A. Lopes, and R. C. Barros, "Self-attention for synopsis-based multi-label movie genre classification," 2018.
[25] J. Wehrmann and R. C. Barros, "Movie genre classification: A multi-label approach based on convolutions through time," *Appl. Soft Comput. J.*, vol. 61, 2017.